\documentclass[10pt,twocolumn,letterpaper]{article}

\usepackage{cvpr}              
\usepackage{times}
\usepackage{epsfig}
\usepackage{graphicx}
\usepackage{amsmath}
\usepackage{amssymb}
\usepackage{mathtools}
\usepackage{bm}
\usepackage{booktabs}
\usepackage{lipsum}
\usepackage{multirow}
\usepackage{tabularx}
\usepackage{makecell}
\usepackage{blindtext}
\usepackage{url}
\usepackage{enumitem}

\usepackage[dvipsnames]{xcolor}

\definecolor{cvprblue}{rgb}{0.21,0.49,0.74}
\usepackage[pagebackref,breaklinks,colorlinks,citecolor=cvprblue]{hyperref}

\title{ConTex-Human: Free-View Rendering of Human from a Single Image with Texture-Consistent Synthesis}

\author{
Xiangjun Gao$^1$ 
\quad Xiaoyu Li$^2$ 
\quad Chaopeng Zhang$^2$
\quad Qi Zhang$^2$ 
\quad Yanpei Cao$^2$ \\
\quad Ying Shan$^2$
\quad Long Quan$^1$\\
\vspace{-15pt}
{$^1$Hong Kong University of Science and Technology\quad $^2$Tencent AI Lab} \\
}

\begin{document}

\twocolumn[{
\renewcommand\twocolumn[1][]{#1}

\maketitle

\centering
\includegraphics[width=\linewidth]{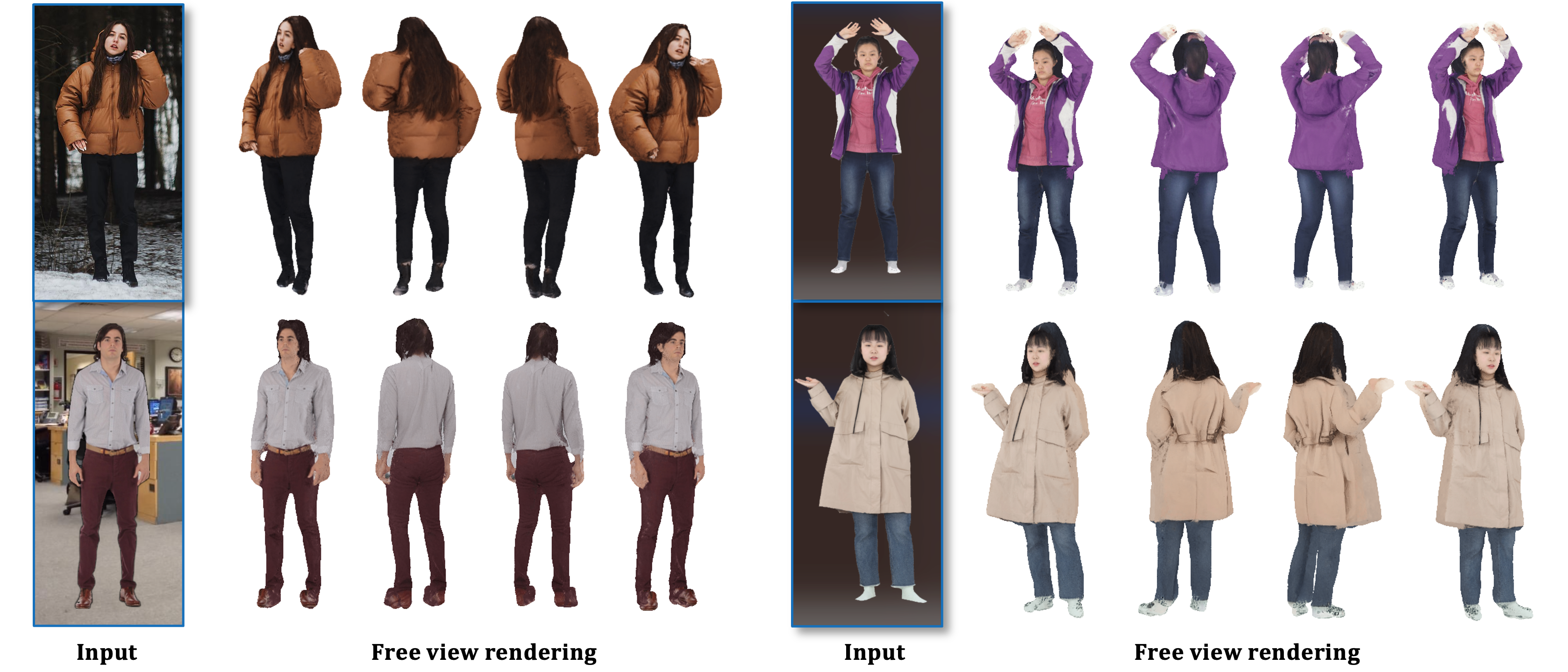}
\vspace{-0.6cm}
\captionsetup{type=figure}
\caption{Our approach \textbf{``ConTex-Human''} can achieve texture-consistent free-view human rendering with high-fidelity using only a single image on different datasets. The left two are from \emph{SSHQ}, right two are from \emph{THuman2.0}.  \textbf{(Check \href{https://gaoxiangjun.github.io/contex\_human}{{\emph{project page}}} for more visual results.)}}
\vspace{0.4cm}
\label{fig:teaser}
}]


\begin{abstract}
In this work, we propose a method to address the challenge of rendering a 3D human from a single image in a free-view manner. Some existing approaches could achieve this by using generalizable pixel-aligned implicit fields to reconstruct a textured mesh of a human or by employing a 2D diffusion model as guidance with the Score Distillation Sampling (SDS) method, to lift the 2D image into 3D space. However, a generalizable implicit field often results in an over-smooth texture field, while the SDS method tends to lead to a texture-inconsistent novel view with the input image. In this paper, we introduce a texture-consistent back view synthesis module that could transfer the reference image content to the back view through depth and text-guided attention injection. Moreover, to alleviate the color distortion that occurs in the side region, we propose a visibility-aware patch consistency regularization for texture mapping and refinement combined with the synthesized back view texture. With the above techniques, we could achieve high-fidelity and texture-consistent human rendering from a single image. Experiments conducted on both real and synthetic data demonstrate the effectiveness of our method and show that our approach outperforms previous baseline methods. 

\end{abstract}    
\vspace{-15pt}
\section{Introduction}
\label{sec:intro}
Free-view human synthesis or rendering is essential for various applications, including virtual reality, electronic games, and movie production. Traditional approaches often require a dense camera rig or depth sensors \cite{collet2015high, dou2016fusion4d} to reconstruct the geometry and refine the texture of the subject, resulting in a tedious and time-consuming process.


Recently, with the advent of implicit fields, remarkable progress has been made in 3D human free-view synthesis from a single RGB image. Several methods~\cite{saito2019pifu, saito2020pifuhd, zheng2021pamir} accomplish this by using pixel-aligned 2D image features as input conditions for subsequent occupancy and color predictions of corresponding 3D points. These methods extract a mesh from the implicit field and predict the vertex colors for free-view rendering. Other works~\cite{kwon2021neural, gao2022mps, hu2023sherf} construct generalizable human neural radiance fields trained on multi-view images, enabling the recovery of 3D humans from a single input image during testing. However, both of these methods tend to produce over-smooth and less fine details due to the smoothness bias of implicit fields and the challenge of inferring the geometry and texture of the entire body from just a single input. To forecast the invisible areas, some methods \cite{qian2023magic123, melas2023realfusion, tang2023make} incorporate a 2D text-to-image diffusion model as guidance and conduct score-distillation-sampling \emph{(SDS)} to optimize a 3D representation from a single image. This approach can also be applied to human images, as shown in our concurrent work TeCH~\cite{huang2023tech}. Nevertheless, the SDS methods tend to produce over-saturation predictions and may not achieve texture-consistent results with the input reference image in the invisible areas, especially for back view images, even when an accurate text prompt is given.

In this paper, we aim to achieve high-fidelity, texture-consistent human free-view rendering using only a single input image, as shown in Figure \ref{fig:teaser}, which presents significant challenges. We propose an innovative framework named ``ConTex-Human''. Under this framework, we decompose our ultimate goal into two key sub-targets. The first one involves generating a texture-consistent back view with fine details. The second sub-target is to paint the side and invisible region with reasonable texture after mapping the input reference and back view onto the reconstructed geometry.

For back view synthesis, we draw inspiration from recent 2D image/video editing methods~\cite{cao2023masactrl, meng2021sdedit, qi2023fatezero, wu2023tune}, which are capable of preserving the style and texture of the original image during the editing process. Our key idea is to query image content from the input reference image to generate a texture-consistent human back view through attention injection, guided by text prompts. However, naively generating the back view using only text prompts would lead to misalignment between the back view image and human geometry. Therefore, we control this process with the depth map as guidance to ensure that the generated back view layout is well-aligned with human geometry. 

In addition to mapping the reference and back view images onto the geometry representation during the optimization, we also need to paint the side region and invisible region. An intuitive solution would be to perform the SDS method on the given person, leveraging the 2D Diffusion model prior. However, only using the SDS loss results in color distortion and over-saturation of texture. To address this problem, we propose a \emph{visibility-aware patch consistency} loss that mitigates the inconsistent side view texture. This approach ensures that the pixel values in the side and invisible regions are close to their neighboring pixels in the front or back regions.

We evaluate our approach on both the synthetic dataset THuman2.0 which has 3D textured scans as ground truth and the real dataset SSHQ which includes people in various poses, clothing, and shapes. Across the experiments, our approach exhibits significant performance in both quantitative and qualitative comparisons. 

In summary,  the contributions of our paper are listed as follows:

\begin{itemize}

    \item We present an innovative framework called ``ConTex-Human'', which could achieve high-fidelity free-view human rendering with consistent texture using single image.

    \item We design a depth and text prompt conditioned back view synthesis module that could maintain texture style and details consistent with the reference image 
    
    \item We proposed a texture mapping and refinement module with a visibility-aware patch consistency loss to synthesize the consistent pixels in invisible areas.

\end{itemize}


\section{Related Work}
\label{sec:relatedwork}

\subsection{Single Image Human Recon. and Rendering}
Reconstructing and Rendering 3D humans from a single image is an ill-conditioned problem that necessitates inferring the geometry or even appearance of the whole body with only one observation. Therefore, a strong prior is usually required to address this issue. Traditionally, parametric body models such as SMPL~\cite{loper2023smpl} are used to estimate the shape and pose from a single image~\cite{guan2009estimating, pavlakos2018learning, bogo2016keep, kanazawa2018end, omran2018neural, lassner2017unite}. 
However, the SMPL mesh representation fails to model complex topologies like dresses and hair. Recently, implicit representations such as NeRF~\cite{mildenhall2021nerf}, occupancy~\cite{mescheder2019occupancy}, and SDF~\cite{park2019deepsdf} have shown impressive results with the ability to model arbitrary topologies and are used to model 3D clothed humans~\cite{saito2019pifu, saito2020pifuhd, xiu2022icon, xiu2022econ, zheng2021pamir, huang2020arch, alldieck2022photorealistic, liao2023high, gao2022mps}. Some of these works focus on the reconstruction of geometry~\cite{saito2020pifuhd, xiu2022icon, xiu2022econ, corona2023structured}, while PIFu~\cite{saito2019pifu}, PAMIR~\cite{zheng2021pamir}, ARCH~\cite{huang2020arch}, PHORHUM~\cite{alldieck2022photorealistic} and S3F~\cite{corona2023structured} would also predict the texture from the image for novel view synthesis. A common limitation of these methods is the requirement of large-scale accurate 3D textured scans as training data. They also tend to predict blurry and over-smooth texture. Recently, with the emergence of large pre-trained models like CLIP~\cite{radford2021learning} and Stable Diffusion~\cite{rombach2022high}, these model priors are also utilized to guide human reconstruction. ELICIT~\cite{huang2023one} utilizes the 3D body shape geometry prior and the visual clothing prior with the CLIP models to create plausible content in the invisible areas of animatable avatar. And TeCH~\cite{huang2023tech} gives the descriptive prompts to the personalized text-to-image diffusion model to learn the invisible appearance through Score Distillation Sampling. Due to the limited expressive ability of text prompts, TeCH suffers from inconsistent texture in the generated areas. In this work, we could reconstruct texture-consistent 3D humans with the help of our texture-consistent back view synthesis method.

\subsection{Image-to-3D Generation}
Recent text-to-image synthesis has achieved high-fidelity generation results benefiting from diffusion models~\citep{sohl2015deep, ho2020denoising} and large aligned image-text datasets. Based on the pre-trained 2D diffusion model, DreamFusion~\cite{poole2022dreamfusion} proposes a Score Distillation Sampling (SDS) method that replaces CLIP models in Dream Fields~\cite{jain2022zero} with the SDS loss to distill 3D models from the text prompt. After that, SDS is widely used in later text-to-3D works~\cite{lin2023magic3d, metzer2023latent, chen2023fantasia3d, wang2023prolificdreamer, zhao2023efficientdreamer, zhu2023hifa}. Inspired by text-to-3D works, image-to-3D that reconstructs 3D models from a single image has also been explored~\cite{xu2023neurallift, melas2023realfusion, deng2023nerdi, tang2023make, qian2023magic123}. Specifically, NeuralLift-360~\cite{xu2023neurallift} derives a prior distillation loss for CLIP-guided diffusion prior to lift a single image to a 3D object. RealFusion~\citep{melas2023realfusion} and NeRDi~\citep{deng2023nerdi} optimize the NeRF representations by minimizing a diffusion loss on novel view renderings with a pre-trained image diffusion model conditioned by a token inverted from input image. Recently, Make-it-3D~\cite{tang2023make} employs a two-stage optimization pipeline that builds textured point clouds to enhance the texture in fine stage, yielding high-quality 3D models according to the given image. Magic123~\cite{qian2023magic123} proposes to use Stable Diffusion~\cite{rombach2022high} as the 2D prior and viewpoint-conditioned diffusion model Zero-1-to-3~\cite{liu2023zero} as the 3D prior simultaneously for SDS loss to generate 3D content from a given image. However, these methods can only address objects with simple geometric structures and textures. For models with more complex geometry and texture, such as 3D humans, obvious artifacts and inconsistent texture would emerge in invisible parts.





\begin{figure*}[htpb]
	\centering
	\includegraphics[width=1.0\linewidth]{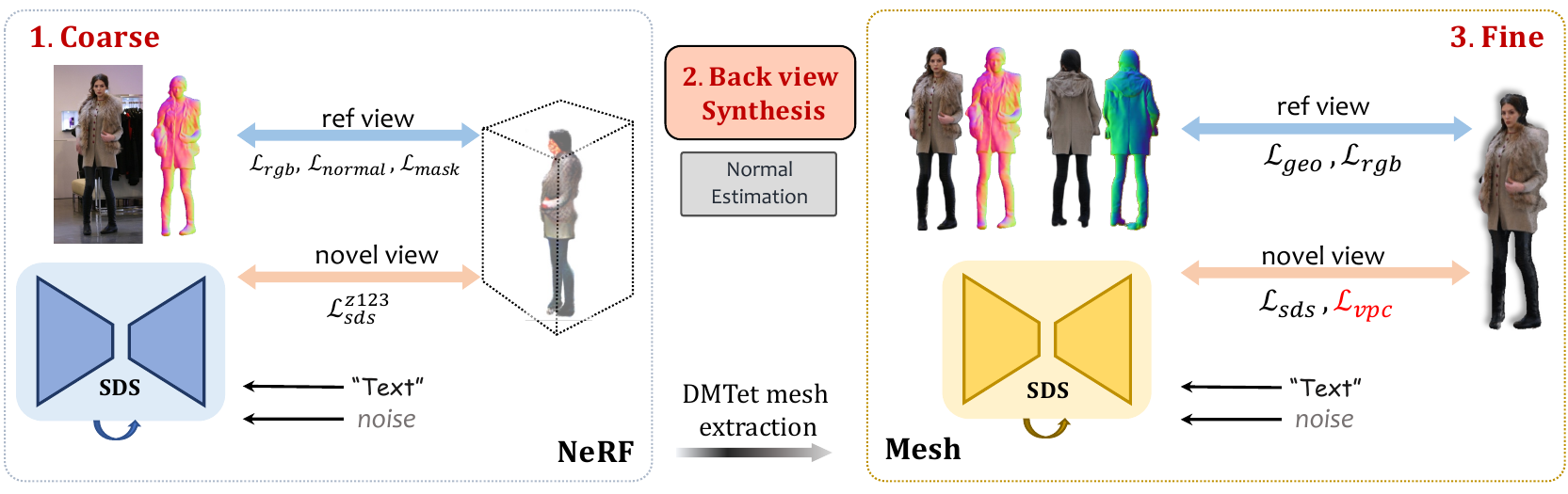}
	\vspace{-15pt}
        \caption{ \textbf{Overview ``ConTex-Human''.} Our framework is composed of three main stages. \textit{(1)Coarse Stage.} Given a human image as reference, we leverage view-aware 2D diffusion model Zero123 to conduct \emph{Score Distillation Sampling (SDS)} to optimize a NeRF. Reference view RGB and normal supervision are also added. \textit{(2)Back view Synthesis Stage.} Coarse image and depth map are utilized to generate texture-consistent and high-fidelity back view. \textit{(3)Fine Stage.} We convert NeRF to DMTet Mesh and optimize mesh with front/back normal map. Texture field is optimized with front/back image, Zero123/Stable-Diffusion \emph{SDS}, and visibility-aware patch consistency loss  $\mathcal{L}_{{vpc}}$.
        }
        \vspace{-5pt}
        \label{method:pipeline}
\end{figure*}

\section{Preliminary}
Recently, DreamFusion \cite{poole2022dreamfusion} has revolutionized the text-to-3D field by proposing the score distillation sampling (SDS) method, which shows that 2D text-to-image diffusion models like Imagen \cite{saharia2022photorealistic} and StableDiffusion \cite{rombach2022high} can be lifted to generate 3D objects base on the text prompts without 3D data. It uses the 2D diffusion model to guide the text-to-3D generation process by optimizing a neural radiance field with a SDS loss that is described as follows:
\begin{equation}
    \begin{aligned}
        \nabla_{\theta}\mathcal{L}_{{sds}}^{sd}
        = \mathbb{E}_{t, \epsilon}\left[w(t)(\hat{\epsilon}_{\phi}(\bm{x}_{t};y,t)-\epsilon)\frac{\partial \bm{x}}{\partial\theta}\right]
    \end{aligned}
    \label{eq:sds}
\end{equation}
where $y$ is the given text prompts, $\bm{x}$ is the rendered image from the 3D representation, $\bm{x}_t$ is the noisy latent after adding Gaussian noise $\epsilon$ to $\bm{x}$, $\hat{\epsilon_{\phi}}$ is the predicted noise, $w(t)$ is the weight function of different noise levels, $\theta$ is the parameters of 3D representations, which is NeRF in DreamFusion.

The \emph{SDS} loss is not only widely used in text-to-3D tasks but also garners significant attention in image-to-3D applications. To learn rich 3D priors from large-scale 3D datasets, such as Objaverse~\cite{deitke2023objaverse}, Zero-1-to-3~\cite{liu2023zero} trains a viewpoint-conditioned diffusion model capable of synthesizing novel views in a feed-forward manner. Given a single image and a target camera pose $\bf{\{R, T\}}$ as input, Zero-1-to-3 can synthesize the corresponding novel view according to the given viewpoint. Furthermore, it can also be employed to guide the optimization of the image-to-3D process using a modified version of the \emph{SDS} loss, formulated as:
\begin{equation}
    \begin{aligned}
        \nabla_{\theta}\mathcal{L}_{{sds}}^{z123}
        = \mathbb{E}_{t, \epsilon}\left[w(t)(\hat{\epsilon}_{\phi}(\bm{x}_{t};\mathrm{\bf{I^r}},t, \bf{R, T})-\epsilon)\frac{\partial \mathrm{\bf I}}{\partial\theta}\right]
    \end{aligned}
    \label{eq:sds}
\end{equation}
where $\bf{I^r}$ is reference image, $\bf I$ is rendered novel view. In this work, we use both of them in different stages.

\section{Method}


Given a single RGB image of a human, our objective is to reconstruct the 3D representation that could render high-fidelity images of the human from various viewpoints. Our approach only requires an RGB image and its foreground mask, which can be easily obtained using an off-the-shelf background removal tool. Our method is illustrated in Figure~\ref{method:pipeline} and consists of three main stages. We first employ the 2D diffusion model to lift the input human image to a radiance field in the coarse stage (\textit{Section 3.1}). Next, we introduce a depth and text-guided attention injection module from the reference to synthesize a texture-consistent image in back view (\textit{Section 3.2}), serving as essential information for the subsequent stage. Finally, we propose a visibility-aware patch consistency loss to reconstruct a 3D mesh for high-quality rendering in the fine stage (\textit{Section 3.3}).

\subsection{Coarse Stage: Radiance Field Reconstruction}
Recent image-to-3D generation methods~\cite{qian2023magic123, tang2023make, melas2023realfusion} that lift a single image into a 3D object often adopt Stable Diffusion (SD) as the diffusion prior. However, we found that the SD guidance frequently results in a tedious optimization process and most importantly, would lead to inconsistent multi-head problems in the optimized 3D object due to data bias in training data of the diffusion model. In the context of 3D humans, this issue can even generate multi-arm and multi-leg geometries for simple poses, let alone handling complex and diverse human poses. To address this problem, we employ the SDS loss based on the Zero-1-to-3 model as the diffusion prior as shown in Eq.2 to optimize an Instant-NGP \cite{muller2022instant} representation.


To optimize the 3D representation, We first employ $\mathcal{L}_{mask}$ between the mask $\bm{\mathrm{M_{r}}}$ extracted from the input image and the rendered mask $\bm{\tilde{\mathrm{M}}_\mathrm{r}}$ in the front view to restrain the human area in 3D space:
\begin{equation}
\mathcal{L}_{{mask}} = \lVert 
        \bm{\mathrm{M_{r}}} - 
        \bm{\tilde{\mathrm{M}}_\mathrm{r}} \rVert_{1}
\end{equation}
In front view, RGB loss is calculated to penalize the difference between the input image $\bm{\mathrm{I_{r}}}$ and rendered results $\bm{\tilde{\mathrm{I}}_\mathrm{r}}$:
\begin{equation}
    \mathcal{L}_{{rgb}} = \lVert 
    \bm{\mathrm{I_{r}}} \odot \bm{\mathrm{M_{r}}} - 
    \bm{\tilde{\mathrm{I}}_\mathrm{r}} \odot \bm{\mathrm{M_{r}}}\rVert_{1}
\end{equation}
In addition, to enforce better geometry and accelerate the training process during optimization, we also incorporate a reference normal constraint for the normal map rendered in the front view. The reference normal is estimated using the normal estimator proposed in ECON~\cite{xiu2023econ}. Therefore, the normal loss between the reference normal $\bm{\mathcal{N}_\mathrm{r}}$ and rendered normal $\bm{\tilde{\mathcal{N}}_\mathrm{r}}$ is formulated as:

\begin{equation}
    \begin{aligned}
        \mathcal{L}_{{normal}} = \lVert 
        \bm{\mathcal{N}_\mathrm{r}} \odot \bm{\mathrm{M_{r}}} - 
        \bm{\tilde{\mathcal{N}}_\mathrm{r}}\odot \bm{\mathrm{M_{r}}} \rVert_{1}
    \end{aligned}
\end{equation}
The overall loss for the coarse stage can be formulated as a
combination of $\mathcal{L}_{sds}^{z123}$, $\mathcal{L}_{mask}$, $\mathcal{L}_{{rgb}}$ and $\mathcal{L}_{{normal}}$. 

\subsection{Texture-Consistent Back View Synthesis}
\label{tex-con}

Although current image-to-3D methods can generate plausible results for invisible areas for the input image, the results tend to be over-saturation, over-smooth, style-inconsistent, and low quality due to the lack of awareness of the input image during synthesizing other areas. Inspired by the recent 2D image editing methods that could maintain content and details of source images during the synthesis and editing process. Our key idea is to query image contents from the input reference image $\bm{\mathrm{I_{r}}}$ and integrate them to synthesize the back view image $\bm{\mathrm{I_{b}}}$ while maintaining the consistent texture details, this process is guided by the text prompt \textbf{T} and depth map \textbf{D}.

Depth map \textbf{D} is able to guide the layout of the $\bm{\mathrm{I_{b}}}$, which is essential for the fine stage to map the texture onto the geometry seamlessly. Text prompt \textbf{T} depicts the human information style such as gender, hair color and style, clothing color and type, etc. Based on the guidance, we propose a depth and text-conditioned texture-consistent back view synthesis module, which utilizes the pre-trained \emph{Depth-Conditioned Stable Diffusion} model and synthesizes a much more highly detailed back view image than the previous methods.

\begin{figure}
	\centering
	\includegraphics[width=1.0\columnwidth]{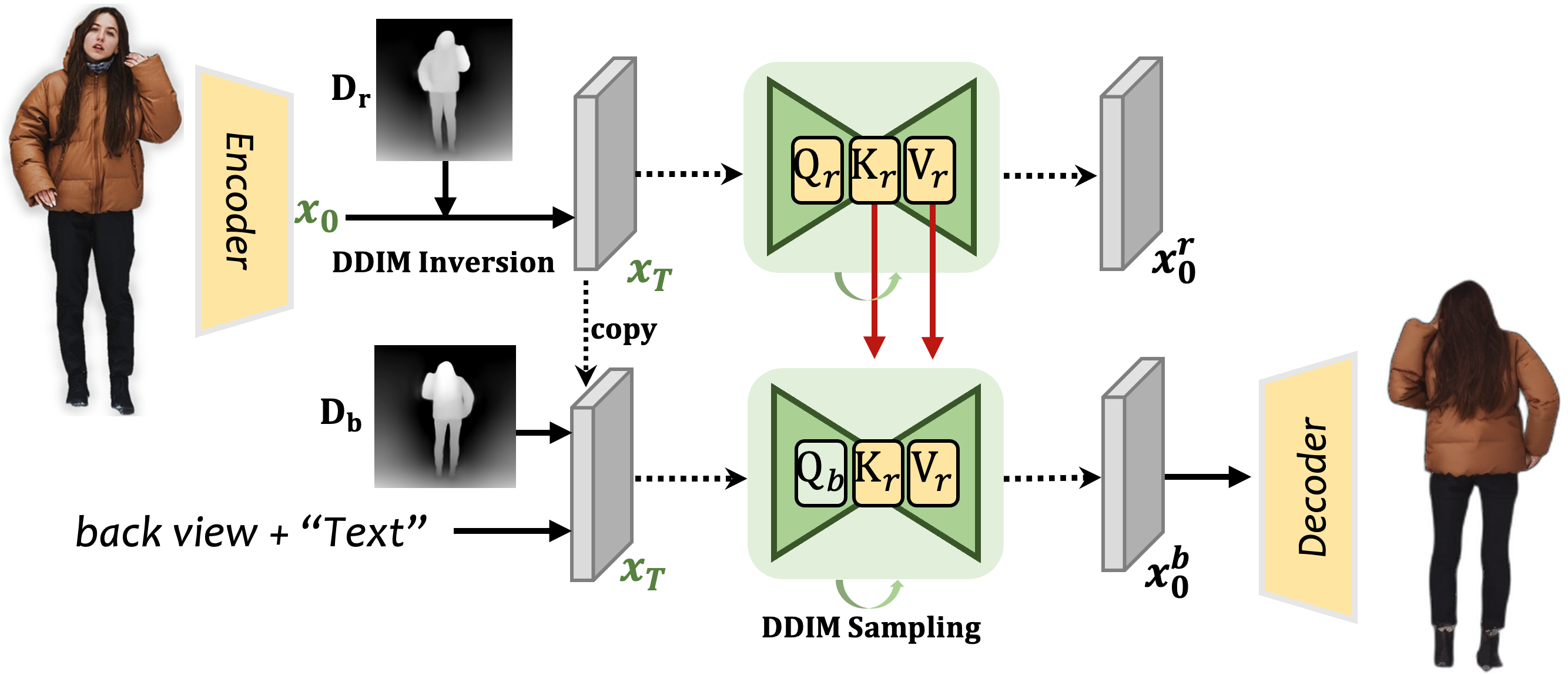}
	\caption{
            \textbf{Illustration of texture-consistent back view synthesis.} We firstly encode the reference image through \emph{SD-Depth} encoder to latent code $\bm x_0$. Then DDIM inversion \cite{song2020denoising} sampling is conducted to get start code $\bm x_T$. The back view is synthesized from $\bm x_T$ through our attention injection method, which is modulated by the depth map.
        }
        \label{fig:backsyn}
	\vspace{-13pt}
\end{figure}

Our texture-consistent back view synthesis module is shown in Figure~\ref{fig:backsyn}. We firstly encode the original front image $\bm{\mathrm{I_{r}}}$ through \emph{SD} encoder to a latent code $\bm x_0$. Then the DDIM inversion \cite{song2020denoising} sampling is conducted on $\bm x_0$ which is concatenated with the front view depth \textbf{D}$\bm_r$ iteratively to get the start noise latent code $\bm x_T$. For the back view synthesis, $\bm x_T$ is copied as the start noise latent code of the back view image, which is concatenated with the back view depth \textbf{D}$\bm_b$ for the subsequent DDIM sampling. Utilizing \textbf{D}$\bm_b$ as conditional information to control the layout, the generated back view is well-aligned with both the coarse stage NeRF and the fine stage Mesh. Moreover, we incorporate the phrase \emph{``back view''} into the original text prompt \textbf{T} to guide the back view synthesis through cross-attention in the \emph{SD} model. 

In addition to the back view synthesis, the front view synthesis process from $\bm x_T$ and \textbf{D}$\bm_r$ is also conducted simultaneously. During the back view and front view synthesis, for the specific time step $t$, we employ an attention injection method to transfer the key feature \textbf{K}$_r$ and value feature \textbf{V}$_r$ in attention layers from the front view branch to the back view branch. In the meanwhile, the back view branch maintains its original query feature \textbf{Q}$_b$ in attention layers. The attention feature transfer is performed iteratively to synthesize the back view. With these proposed operations, the detailed texture from the front view image can be transferred to the back view, simultaneously, maintaining the back view depth layout that is view-consistent with the front view geometry and being well aligned in accordance with the original text description.

\subsection{Fine Stage: High-Fidelity Mesh Reconstruction}
The coarse stage generates only a rough geometry and low-quality texture, represented by a density field and a color field. Therefore, we introduce a fine stage to refine the geometry and texture from the coarse stage by utilizing the content details in the reference image and generated the back view image from our method. Compared to recent works~\cite{tang2023make, qian2023magic123, huang2023tech} that employ the SDS loss to optimize the full texture which could suffer from over-saturation, blurriness, over-smoothing, and style inconsistency in the generated areas. Our solution, which maps the generated back view that is style and texture-consistent with the front view onto the refined geometry, combining the proposed visibility-aware patch consistency regularization, achieves more 3D-consistent and high-fidelity results.

\subsubsection{Geometry Reconstruction}
We adopt DMTet\cite{shen2021deep}, a hybrid SDF-Mesh representation, for the reconstruction in the fine stage, which is capable of generating high-resolution 3D shapes and allows for efficient differentiable rendering. To initialize DMTet, we set the SDF value of each vertex $v_i$ using the density field from the coarse stage, and the deformation vector $\triangle v_i$ is set to $\bm0$. During geometry optimization, a triangle mesh is extracted from DMTet. We employ a differential rasterizer \cite{Laine2020diffrast} to render the normal map from a given viewpoint. 

To regularize the geometry during optimization, we also employ a normal constraint as in the coarse stage. One straightforward way is to estimate the normal maps of both the front and back view from the reference image as the supervision using an existing normal estimator in ECON~\cite{xiu2023econ}. However, alignment issues arise between the estimated back view normal and the reconstructed geometry due to the different camera settings. In view of this,  we alternatively adopt the estimated back normal $\bm{\mathcal{N}_\mathrm{b}}$ from the synthesized texture-consistent back view that is generated in Section~\ref{tex-con}. This normal is well-aligned with our initialized geometry and synthesized back view.


Given that the reference view normal and back view normal encompass most of the human region, a reasonable transition between the reference and back views can be achieved after applying mesh normal smoothness and laplacian smoothness constraints. Finally, the geometry reconstruction loss in the fine stage can be written as follows:
$$
\mathcal{L}_{geo} = \lVert \bm{\mathcal{N}_\mathrm{r}} - \bm{\tilde\mathcal{N}_\mathrm{r}} \rVert_2  + \lVert \bm{\mathcal{N}_\mathrm{b}} - \bm{\tilde\mathcal{N}_\mathrm{b}} \rVert_2 + \mathcal{L}_{smooth}
$$
where $\bm{\tilde\mathcal{N}_\mathrm{r}}$ and $\bm{\tilde\mathcal{N}_\mathrm{b}}$ are the rendered mesh normal maps using Nvdiffrast~\cite{Laine2020diffrast}, respectively. $\bm{\mathcal{N}_\mathrm{r}}$ and $\bm{\mathcal{N}_\mathrm{b}}$ are the reference and back view ground truth normal maps, respectively, estimated using ECON~\cite{xiu2023econ} normal estimator for the reference image and generated back view image. 

\subsubsection{Texture Mapping and Refinement}
After the geometry reconstruction in the fine stage, the next step is to generate texture by mapping the reference front image $\bm{\mathrm{I_{r}}}$ and the generated back view image $\bm{\mathrm{I_{b}}}$ to the refined geometry. Similar to \cite{lin2023magic3d}, we adopt an Instant-NGP to represent the 3D texture field. For each pixel $\bm x_i$ in the sampled image, we first calculate its ray-mesh intersection's 3D position $\bm p_i$. Then a latent feature is interpolated from the Instant-NGP feature grid and is fed to a tiny layer MLP network to decode a color value. The texture field is first regularized using the front reference image $\bm{\mathrm{I_{r}}}$ and the generated back view image $\bm{\mathrm{I_{b}}}$ as the supervision:
\begin{equation}
\mathcal{L}_{rgb} = \lVert \bm{\mathrm{I}_\mathrm{r}} - \bm{\tilde\mathrm{I}_\mathrm{r}} \rVert_2  + \lVert \bm{\mathrm{I}_\mathrm{b}} - \bm{\tilde\mathrm{I}_\mathrm{b}} \rVert_2
\end{equation}
where $\bm{\tilde\mathrm{I}_\mathrm{r}}$ and $\bm{\tilde\mathrm{I}_\mathrm{b}}$ are the rendered front and back images from the texture field, respectively.

\begin{figure}[htpb]
	\centering
	\includegraphics[width=1.0\columnwidth]{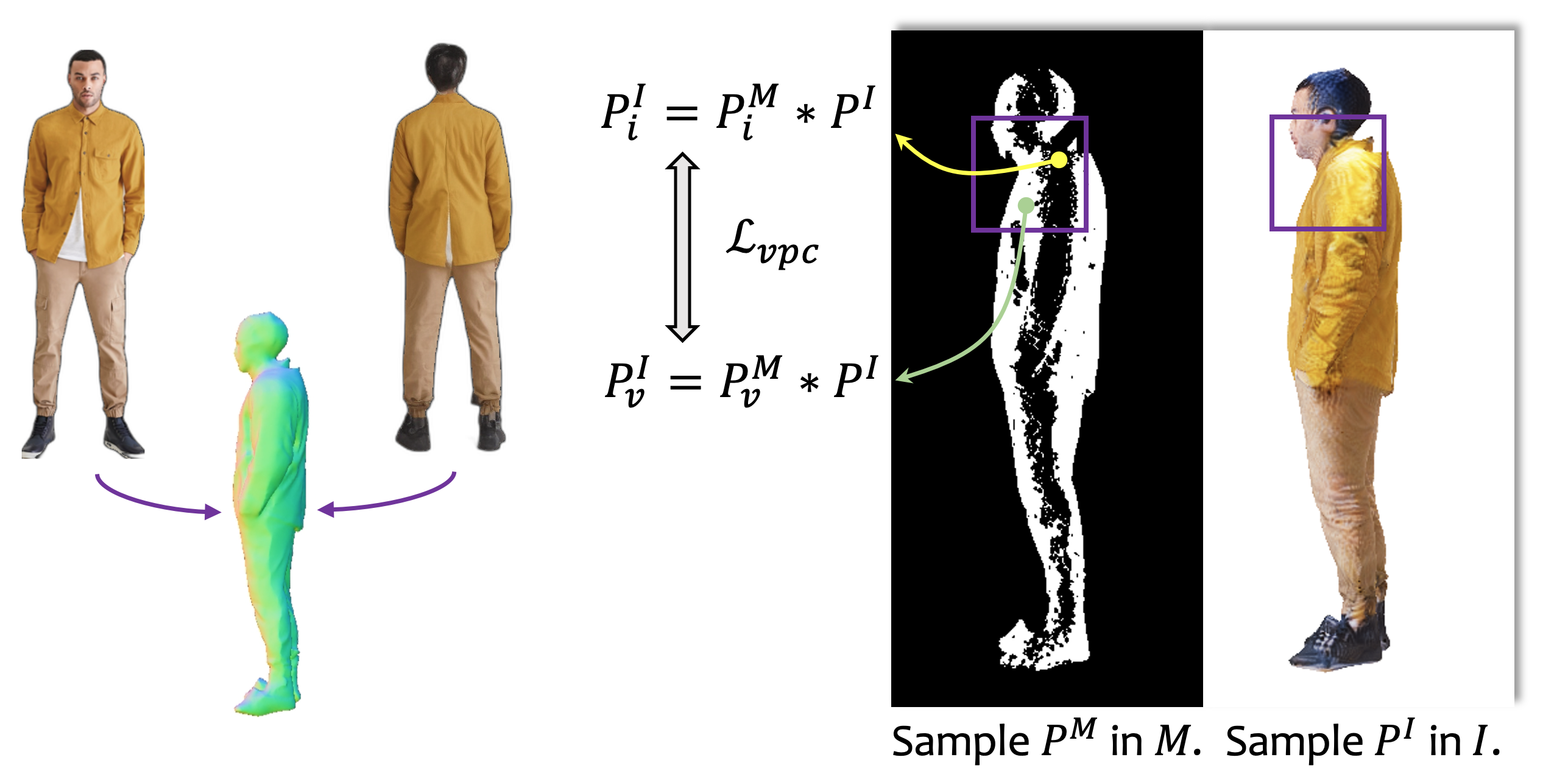}
        \vspace{-15pt}
	\caption{
            \textbf{Illustration of visibility-aware patch consistency loss $\mathcal{L}_{vpc}$}. After mapping the front/back view to the geometry, we conduct \emph{SDS} to complete side and invisible region. To remove the color distortion caused by \emph{SDS}, we sample random patches \textit{P}$^I$ which are divided into visible region \textit{P}$_v^I$ and invisible region \textit{P}$_i^I$. The $\mathcal{L}_{vpc}$ is calculated between regions. 
        }
        \label{fig:chamf}
\end{figure}

Although the front and back view images could cover most of the texture for the human, there are still some missing textures in the side view and self-occluded region. To complete the missing texture, similar to ~\cite{qian2023magic123}, we adopt a SDS combination of both Stable-Diffusion and Zero-1-to-3 models to optimize the texture field:

\begin{equation}
    \begin{aligned}
        \mathcal{L}_{{sds}}
        = \lambda_1 \mathcal{L}_{{sds}}^{sd} + \lambda_2 \mathcal{L}_{{sds}}^{z123}
    \end{aligned}
    \label{eq:sds}
\end{equation}
The combined SDS loss can fill the missing region guided by text prompts. However, in many cases, there is an obvious texture transition and inconsistent style in the filled part between the front and back view images. To address this problem, we propose a \emph{visibility-aware patch consistency loss} for refinement, which could alleviate the inconsistent side view texture as shown in Figure \ref{fig:chamf}. To be specific, for each pixel in the front view image and back view image, we find its intersection with the corresponding mesh triangle face through rasterization. The vertices on the face closest to the intersection are set to 1, indicating that they are visible to $\bm{\mathrm{I_\mathrm{r}}}$ or $\bm{\mathrm{I_\mathrm{b}}}$. Vertices that are invisible to $\bm{\mathrm{I_\mathrm{r}}}$ and $\bm{\mathrm{I_\mathrm{b}}}$ are set to 0. 

Our key insight is that the pixels in the invisible region should have a consistent color with their neighbor visible pixels within a patch. To achieve this, we first sample a random viewpoint in camera space and render an RGB image  $\bm{{I}}$ and its visibility map \textbf{\textit{M}}. Then we sample a random patch \textit{P}$^I$ in $\bm{{I}}$ and its visibility map \textit{P}$^M$ in \textbf{\textit{M}}. In this patch, the invisible pixels \textit{P}$^I_i$ can be calculated by (\textit{P}$^I_i$ = \textit{P}$^I$ * \textit{P}$_i^M$), and the visible pixels \textit{P}$^I_v$ can be calculated by (\textit{P}$^I_v$ = \textit{P}$^I$ * \textit{P}$_v^M$). Then we calculate the visibility-aware patch consistency loss as follows:

\vspace{-5pt}
\begin{equation}
    \begin{aligned}
        \mathcal{L}_{{vpc}}
        = \sum_{p \in \textit{P}^I_i} \min_{q \in \textit{P}^I_v} \|p - q\|^2
    \end{aligned}
    \label{eq:chf}
\end{equation}

The overall loss for the texture mapping and refinement can be formulated as a combination of $\mathcal{L}_{rgb}$, $\mathcal{L}_{sds}$ and $\mathcal{L}_{vpc}$. Due to space constraints, we will provide more implementation details in the \textit{supplementary materials}.

\section{Experiments}


\subsection{Datasets}
We describe the human datasets used in the experiments, including a synthetic dataset rendered by 3D textured scans THuman2.0~\cite{yu2021function4d},  and a real dataset with high-quality full-body human images, SSHQ~\cite{fu2022stylegan}.

 \textbf{THuman2.0} is a 3D human model dataset that contains 500 high-quality human scans captured by a dense DSLR rig. For each scan, it provides a 3D model along with the corresponding texture map. In all 500 static scans, the same person might appear multiple times with different poses and clothes. As our method is person-specific, evaluating all the data would be a tedious process. Therefore, we selected 30 subjects with different identities, poses, and clothes for evaluation. For each subject, we used the front view as the input for our method. To evaluate our method, we rendered ten views that surround the center human as ground truth novel views using PyTorch3D. We rendered images at a resolution of 648×648 pixels, where the height of the human region comprises approximately 70\% of the image, resulting in a height of around 455 pixels.

 \textbf{SSHQ} is a dataset consisting of high-quality full-body human images at a resolution of 1024×512. SSHQ covers a wide range of races, clothing styles, and poses. Similar to THuman2.0, we selected 30 subjects with only a single image for evaluation. We remove the background, resize the image to a resolution of 648×648 pixels, and reposition the human to occupy approximately 70\% of the image's height, ensuring it remains centered.


\subsection{Metrics and Methods for Comparison}
To evaluate our methods, we compare our method with PIFu~\cite{saito2019pifu} and PaMIR~\cite{zheng2021pamir} which are single image human reconstruction methods including the texture that can be inferred in a feed-forward manner. Additionally, we also compare our results with a recent image-to-3D generation method Magic123~\cite{qian2023magic123} which is a per-subject optimization method like ours. For the THuman2.0 dataset, since both the input view and novel views have the ground truth, we evaluate all methods on the rendered images in input view and novel views using commonly used metrics: PSNR, SSIM, LPIPS, and CLIP. For the SSHQ dataset that only has the input image for evaluation, following the experiments in Magic123, we adopt CLIP similarity to measure the consistency between the input image and the rendered novel view images and use LPIPS to measure the accuracy between the input image and the rendered reference view.

For TeCH~\cite{huang2023tech}, a concurrent work to our method, released their code two weeks before the paper submission. However, it still exhibits inconsistencies in the generated areas. We provide a comparison with it in our \emph{suppl.} 

\begin{figure*}[htpb]
	\centering
	\includegraphics[width=1.0\linewidth]{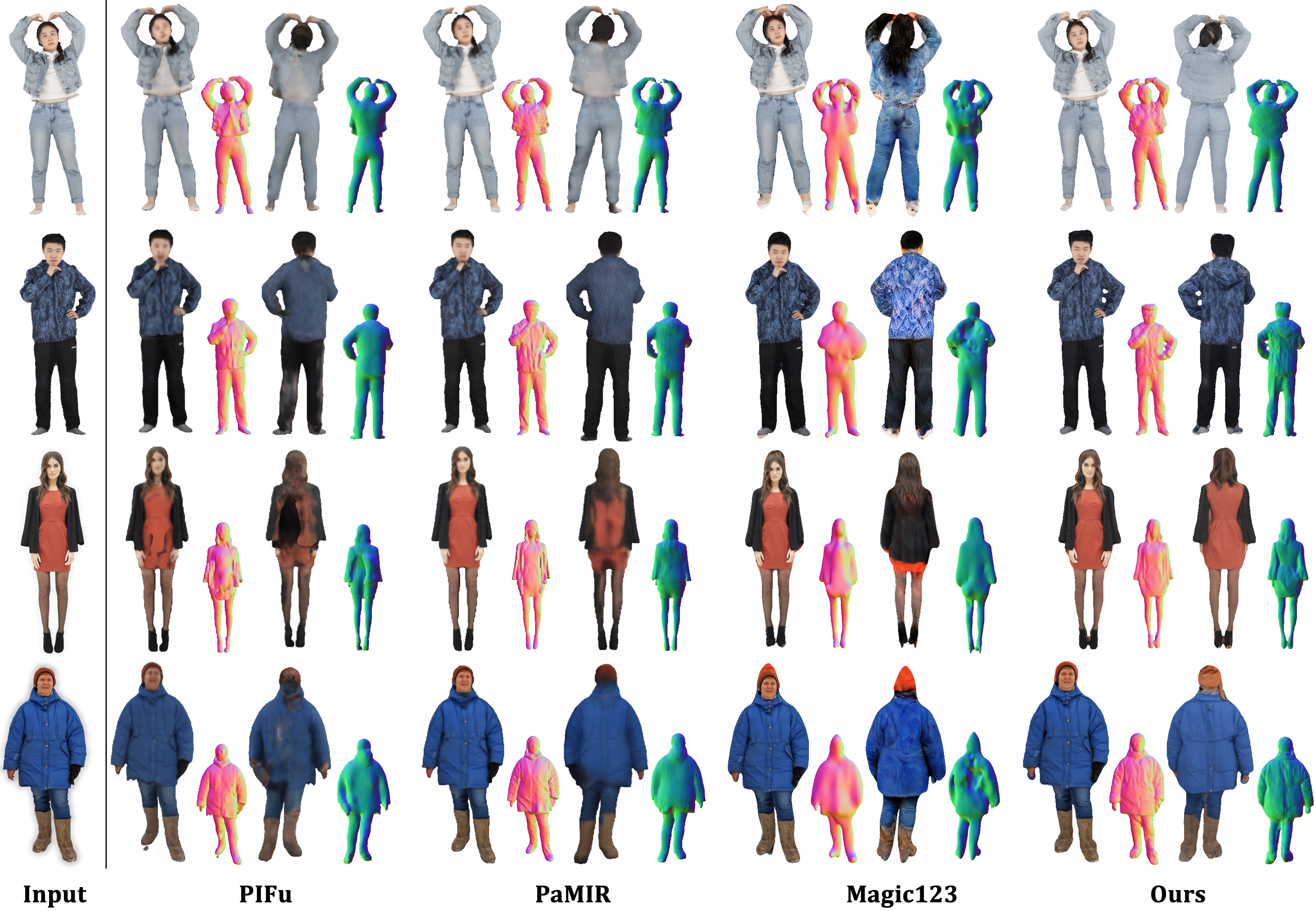}
	\vspace{-16pt}
        \caption{{\textbf{Qualitative results on THuman2.0 and SSHQ dataset.} Row 1\&2 are THuman2.0 samples, Row 3\&4 are SSHQ samples. PIFu and PaMIR tend to predict blurred rendering results, especially in the back view. Magic123 has difficulty in predicting consistent texture. Compared with them, our methods accurately render texture-consistent and high-fidelity novel views, please \textbf{\textit{Zoom in}}  for the details.}}
        \label{exp:comp}
	\vspace{-10pt}
\end{figure*}

\subsection{Evaluation}
 \textbf{Results on the THuman2.0 dataset.}
 Table~\ref{tab:comp} summarizes the quantitative comparison of our proposed method with the baseline methods on THuman2.0 dataset. PIFu and PaMIR sample dense 3D points in 3D space and the sampling points are projected onto the input image plane to get the interpolated image features for occupancy prediction. Then an explicit mesh is extracted using the Marching-Cube\cite{lorensen1998marching} algorithm. For rendering, they predict a color value for each mesh vertex. Similar to us, Magic123 is a per-subject optimization method that reconstructs the given subject by lifting the 2D diffusion models to 3D using the SDS loss. In Table~\ref{tab:comp}, we demonstrate that our approach surpasses all the baseline methods with respect to PSNR, LPIPS, and CLIP metrics while achieving the second-best performance in terms of SSIM. This indicates that our results are more closely aligned with the ground truth.

In row 1\&2 of Figure \ref{exp:comp}, we present qualitative results of our method and baseline methods in the front view and back view rendered images. Our method produces a clearer, more detailed, and more photo-realistic front view and back view than PIFu and PaMIR. Magic123 effectively preserves the details of the input image by employing an RGB loss in the input view. However, it generates over-saturated and texture-inconsistent results in the back view due to the SDS loss. In contrast, our results could produce a more realistic result and consistent texture than Magic123 in different views. Our method can also handle complicated textures as the jacket shown in row 2 of Figure~\ref{exp:comp}. For more visual results, please refer to our \textit{supplementary materials}.

\begin{table}[t!]
\centering
\small
\caption{\textbf{Quantitative comparison of our method} with PIFu, PaMIR, Magic123 on THuman2.0 and SSHQ datasets in terms of SSIM, PSNR, LPIPS, and CLIP. ($\uparrow$ means higher is better, $\downarrow$ means lower is better.) }
\label{tab:comp}
\setlength{\tabcolsep}{0.6mm}{
    \begin{tabular}{ccccc|cc}
        \toprule
        \multicolumn{1}{c}{\multirow{2}{*}{Method}} & \multicolumn{4}{c}{THuman2.0} & \multicolumn{2}{c}{SSHQ} \\
        \cmidrule{2-7}  & SSIM $\uparrow$ & PSNR$\uparrow$ & LPIPS$\downarrow$  & CLIP$\uparrow$   & LPIPS$\downarrow$ & CLIP$\uparrow$ \\
        \midrule
PIFu     & 0.921 & 20.4 & 0.079 & 0.889  & 0.068 & 0.873 \\
PaMIR    & \textbf{0.925} & 21.0 & 0.072 & 0.913  & 0.064 & 0.887 \\
Magic123 & 0.903 & 18.8 & 0.099 & 0.910  & \textbf{0.056} & 0.882 \\
Ours     & {0.923} & \textbf{21.4} & \textbf{0.063} & \textbf{0.932}  & 0.059 & \textbf{0.903} \\
        \bottomrule
    \end{tabular}%
    }	
\vspace{-10pt}
\end{table}

 \textbf{Results on the SSHQ dataset.}
In addition to evaluating the performance on the rendered dataset, we also evaluate all methods using the real images from the SSHQ dataset. As there is no ground truth for the novel view, we compute the LPIPS metric in the front view and the CLIP similarity in the novel view. The LPIPS metric measures how closely the rendered front view matches the input image, while the CLIP similarity evaluates the resemblance between the rendered novel view and the input image. Table~\ref{exp:comp} shows the quantitative results of our method and the baseline methods. We are the second-best in LPIPS in the front view which is marginally lower than Magic123, but we are the best in the CLIP similarity, showing that our rendered novel views are closer to the input image.

In rows 3 and 4 of Figure \ref{exp:comp}, we present a qualitative comparison, which shows that our model is capable of producing more detailed front and back views than PIFu and PaMIR, as well as much more consistent texture with the input image than Magic123. Furthermore, our model is even able to handle various clothing types, such as loose coats, boots, dresses, and hats, as shown in Figure \ref{exp:comp}. Although our primary focus is not on geometry, we demonstrate an improved back view normal map with enhanced details.



\begin{table}[t!]
\centering
\small
\caption{Comparison between \textbf{no back view synthesis} (w/o back), \textbf{no visibility-aware patch consistency loss} (w/o VPC), and \textbf{full model} on THuman2.0 dataset.  ($\uparrow$ means higher is better, $\downarrow$ means lower is better.)}
 \label{tab:ab}
\setlength{\tabcolsep}{0.6 mm}{
    \begin{tabular}{ccccc}
        \toprule
        \multicolumn{1}{c}{\multirow{2}{*}{Method}} & \multicolumn{4}{c}{THuman2} \\
        \cmidrule{2-5} & SSIM$\uparrow$ & PSNR$\uparrow$ & LPIPS$\downarrow$ & CLIP$\uparrow$   \\
        \midrule
w/o back \& w/o VPC  & 0.9126 & 20.188 & 0.9051 & 0.0778 \\
with back \& w/o VPC & 0.9220 & 21.27  & 0.9328 & 0.0658 \\
Full model           & \textbf{0.9231} & \textbf{21.35}  & \textbf{0.9315} & \textbf{0.0634} \\
        \bottomrule
    \end{tabular}%
    }	
    \vspace{-5pt}
\end{table}

\subsection{Ablation Study}

We conduct ablation studies to analyze how the texture-consistent back view synthesis module and the visibility-aware patch consistency loss in the texture mapping and refinement module affect the performance of our methods. These ablation studies for quantitative results are conducted on the THuman2.0 dataset, as it provides the ground truth data in novel views. For the qualitative results, we test the cases in both the THuman2.0 and SSHQ datasets.

\begin{figure}
	\centering
	\includegraphics[width=1.0\columnwidth]{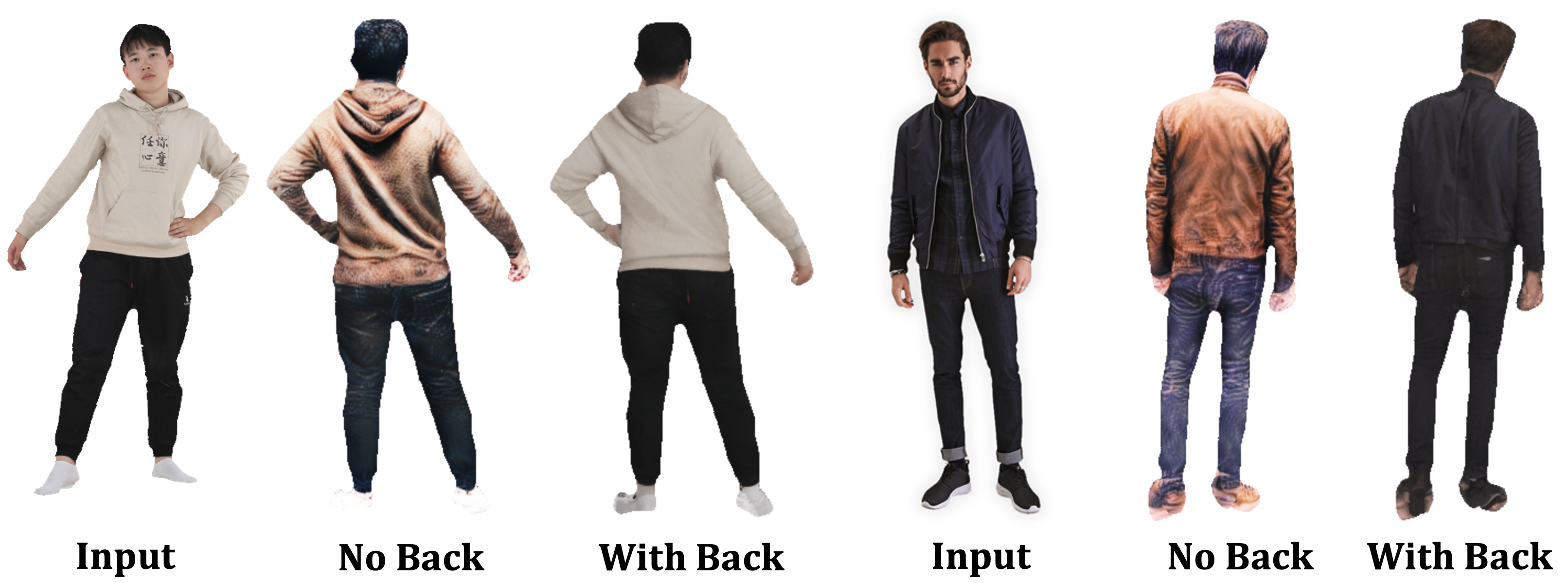}
	
        \caption{
            \textbf{Qualitative results between \emph{no texture-consistent back view synthesis} module} and \textbf{\emph{full model}} on THuman2.0 and SSHQ.
        }
        \label{fig:ab_back}
	\vspace{-10pt}
\end{figure}

\textbf{Texture-consistent back view synthesis.} To validate the importance of using a texture-consistent back view synthesis, we remove the synthesized back view image during the texture optimization in the fine stage, which means we remove the back view regularization in $\mathcal{L}_{rgb}$ and the visibility-aware patch consistency loss $\mathcal{L}_{vpc}$ that also relies on the back view image. In this setup, our method is more similar to previous image-to-3D methods. As shown in Table \ref{tab:ab}, the performance drops significantly when the texture-consistent back view image is removed, indicating that this design is critical for the single image 3D human free-view rendering. The visual comparison is shown in Figure \ref{fig:ab_back}. It can be observed that, without the texture-consistent back view, the textures tend to be of significantly lower quality and, most importantly, lack consistency with the input view.

\begin{figure}
	\centering
	\includegraphics[width=1.0\columnwidth]{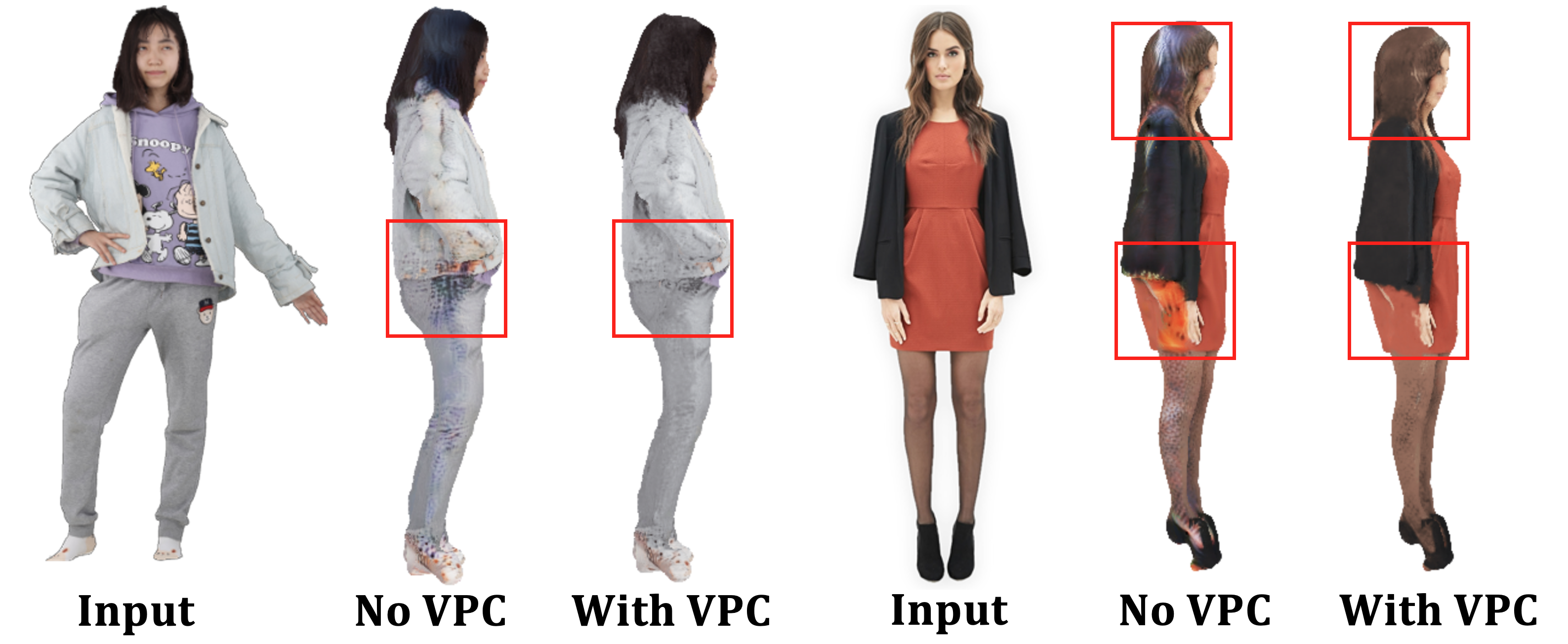}
	\vspace{-15pt}
        \caption{
            \textbf{Qualitative results between \emph{no visibility-aware patch consistency loss}} and \textbf{\emph{full model}} on THuman2.0 and SSHQ. 
        }
        
        \label{fig:ab_vpc}
        \vspace{-10pt}
	
\end{figure}

\textbf{Visibility-aware patch consistency loss.}
Additionally, we validate the effectiveness of the proposed visibility-aware patch consistency loss (VPC) by excluding it from the optimization process. The performance also decreases as shown in Table \ref{tab:ab}. The visual examples presented in Figure \ref{fig:ab_vpc} show that without the VPC loss, severe color distortion, as well as color inconsistency, would appear in the side region. We attribute this inconsistency to the inability of the SDS optimization to guide the model towards an optimal convergence solution that best satisfies the front view.





\vspace{-5pt}
\section{Limitations}
\vspace{-5pt}
The depth and text-conditioned back view synthesis, along with visibility-aware patch consistency loss, enable us to achieve remarkable free-view human rendering with consistent texture. However, there are a few limitations. 1) We are unable to generate very impressive high-quality geometry. The resulting mesh exhibits coarse geometry in the hand and foot regions. Besides, 
if the coarse stage produces geometry with concave areas or significantly incorrect poses, the fine-stage mesh refinement cannot adequately compensate for these errors. 2) Although the side and invisible regions exhibit color-consistent predictions, their quality is not as high as that of the front and back views, and they occasionally contain some noise. 3) Similar to NeRF, our proposed method is trained in a person-specific setting, which requires over one hour to achieve training.
\vspace{-5pt}

\section{Conclusion}
\vspace{-5pt}

In this paper, we introduced a novel framework for single image free-view 3D human rendering. We proposed a module for texture-consistent and high-fidelity back view synthesis, which is well-aligned with the input reference image. The texture mapping module with visibility-aware patch consistency loss is proposed for side and invisible region inpainting. Experiments on the \emph{THuman2.0} and \emph{SSHQ} demonstrated that the proposed model achieves state-of-the-art performances on free-view image synthesis.

\newpage
{
    \small
    \bibliographystyle{ieeenat_fullname}
    \bibliography{main}
}

\clearpage
\setcounter{page}{1}
\maketitlesupplementary

\textbf{Overview.} The supplementary material has the following contents:

\begin{itemize}[nosep,left=1.5em]
    \item Coarse stage implementation details
    \item Fine stage implementation details
    \item Compare with TeCH
    \item More visual comparison
    \item Visual results video demo
\end{itemize}

\section*{A. Coarse Stage Details} 

\noindent \textbf{Pre-Process} Given a single image of a specific person, we first adopt the off-the-shelf background removal tool from \url{https://github.com/danielgatis/rembg} to attain the human foreground mask $\mathbf{M}$. Based on the foreground mask $\mathbf{M}$, we create an RGBA image with 648×648 resolution and make sure that the valid human region occupies approximately 70 \% of the image’s height, ensuring it remains centered.

Besides the mask, we also need the reference image normal map. In practice, we employ the designed normal estimator \textbf{N} from ECON\cite{xiu2023econ}. Note that, \textbf{N} is conditioned with an optimized SMPL normal map. Therefore, our optimized geometry also incorporates the human pose information.

\vspace{5pt}
\noindent \textbf{Camera Setting.} For the coarse stage, we optimize the neural radiance field(NeRF)\cite{mildenhall2021nerf} with 128×128 resolution. The goal of the coarse stage is to supply a coarse geometry with a roughly accurate human pose and boundary for \emph{back view synthesis stage}  and \emph{fine stage}. The elevation and azimuth degree of the reference image is set to \textbf{0.} as default. 

For the camera setting during \emph{Score Distillation Sampling}, the elevation range is set to [-30$^{\circ}$, 60$^{\circ}$], and the azimuth range is set to [-180$^{\circ}$, 180$^{\circ}$]. The camera distance is set to 3.8 as default, camera field of view (FOV) is set to 20$^{\circ}$ which is aligned with Zero-1-to-3\cite{liu2023zero}.

\vspace{5pt}
\noindent \textbf{3D Representation.} We employ a multi-resolution hash grid from Instant-NGP\cite{muller2022instant} as the 3D NeRF representation. We use 16 levels of hash dictionaries of size $2^{19}$, each entry is with a dimension 2 feature vector. The 3D grid resolution range from $2^{4}$ to $2^{12}$ with an exponential growth rate of 1.447. A two-layer tiny MLP with 64 hidden units is adopted to decode the concatenated features interpolated from Instant-NGP to RGB color and volume density. The background is a ``white'' solid color background. We sample 512 points along each ray.

\vspace{5pt}
\noindent \textbf{Score Distillation Sampling.} We sample images with a batch\_size of 4 each iteration for \emph{Score Distillation Sampling.} We sample the timestep $t \sim \mathcal{U}(0.2, 0.6)$, the classifier-free guidance weight is set to 5. 

The overall $\mathcal{L}_{coarse}$ loss for the coarse stage can be formulated as a
combination of $\mathcal{L}_{sds}^{z123}$, $\mathcal{L}_{mask}$, $\mathcal{L}_{{rgb}}$ and $\mathcal{L}_{{normal}}$:
\vspace{-3pt}
\begin{equation}
    \begin{aligned}
        \mathcal{L}_{coarse} = \lambda_1 \mathcal{L}_{sds}^{z123} + \lambda_2 \mathcal{L}_{{rgb}} + \lambda_3 \mathcal{L}_{{normal}} + \lambda_4 \mathcal{L}_{mask}
    \end{aligned}
\end{equation}

where in practice $\lambda_1=1.0$, $\lambda_2=1000$, $\lambda_3=1000$, $\lambda_4=1000$, some additional constrain like density sparsity and normal smoothness are also employed during optimization. We optimize the coarse stage using Adam optimizer for 3000 steps with a learning rate 5×10$^{-3}$.

\section*{B. Fine Stage Details}
\noindent \textbf{Geometry Optimization.} We adopt DMTet\cite{shen2021deep} in the fine stage, a hybrid SDF-Mesh representation, the DMtet resolution is set to 256×256×256. 

The overall $\mathcal{L}_{fine}^{geo}$ loss for the coarse stage can be formulated as a
combination of $\mathcal{L}_{sds}^{z123}$, $\mathcal{L}_{mask}$, $\mathcal{L}_{{rgb}}$ and $\mathcal{L}_{{normal}}$:
\vspace{-3pt}
\begin{equation}
    \begin{aligned}
        \mathcal{L}_{fine}^{geo} = \lambda_1 \mathcal{L}_{normal} + \lambda_2 \mathcal{L}_{mask} + \lambda_3 \mathcal{L}_{{lap}} + \lambda_4 \mathcal{L}_{smooth}
    \end{aligned}
\end{equation}

where $\mathcal{L}_{smooth}$ is the mesh normal constraint, $\mathcal{L}_{lap}$ is the mesh laplacian constraint. In practice $\lambda_1=10000$, $\lambda_2=50000$, $\lambda_3=1000$, and $\lambda_4=1000$. We optimize the geometry stage using Adam optimizer for 3000 steps with a learning rate 1×10$^{-2}$. In steps 2000$\sim$3000 step, $\lambda_3$ and $\lambda_4$ are set to 100 for more human geometry details.

\vspace{5pt}
\noindent \textbf{Texture Field.} 
We employ another multi-resolution hash grid to represent the texture field. We use 14 levels of hash dictionaries of size $2^{19}$, each entry is with a dimension 2 feature vector. Same as the coarse stage, the 3D grid resolution ranges from $2^{4}$ to $2^{12}$. A two-layer tiny MLP with 64 hidden units is adopted to decode the concatenated features to RGB color. The background is a ``white'' solid color background.

\vspace{5pt}
\noindent \textbf{Camera Setting.} The camera setup of the fine stage is similar to the coarse stage except that the elevation degree range is [-45$^{\circ}$, 45$^{\circ}$] and the image resolution is 648×648.

\begin{figure*}[htpb]
	\centering
	\includegraphics[width=1.0\linewidth]{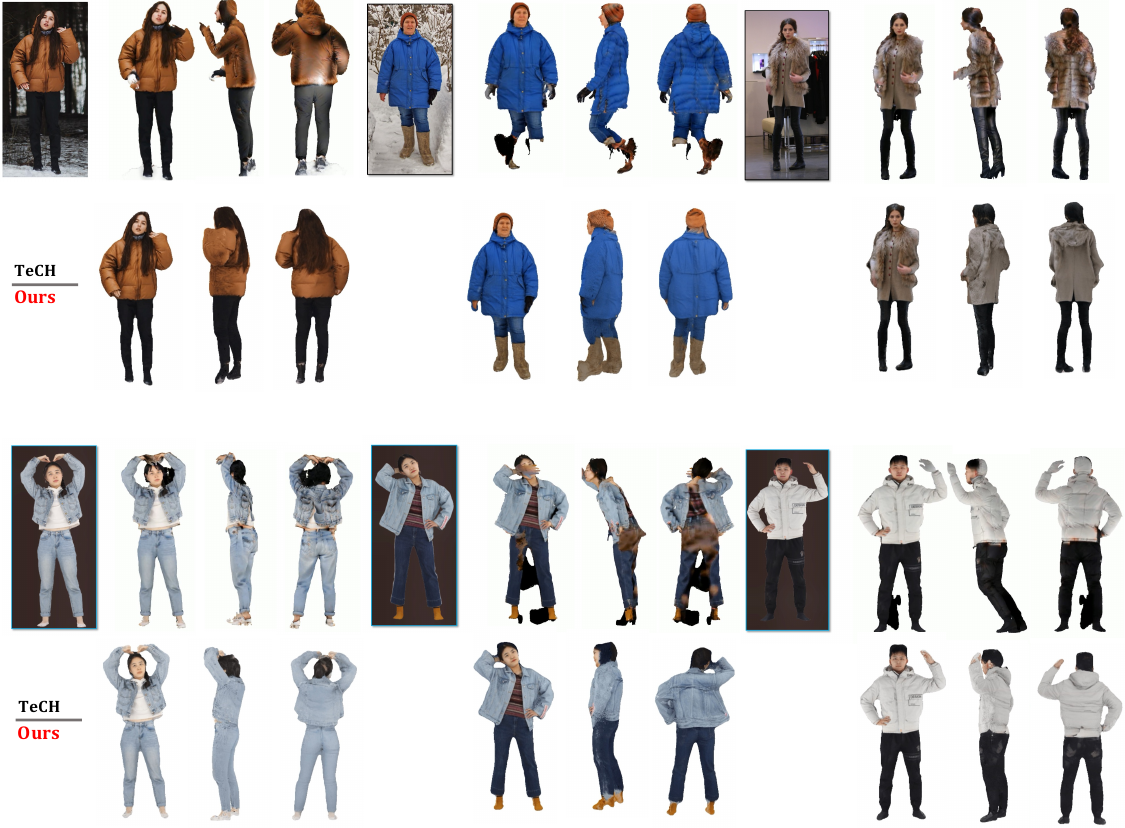}
	\vspace{-16pt}
        \caption{{\textbf{Qualitative comparison with TeCH\cite{huang2023tech}results on THuman2.0 and SSHQ dataset.} Compared with TeCH, our methods have a consistent texture with input images. Row 1\&3 are TeCH results, Row 2\&4 are our results. Please \textbf{\textit{Zoom in}}  for the details.}}
        \label{exp:comp_tech}
	\vspace{-10pt}
\end{figure*}

\vspace{5pt}
\noindent \textbf{Score Distillation Sampling.}  We sample images with a batch\_size of 1 for each iteration for SDS. For Zero-1-to-3 SDS, we sample the timestep $t \sim \mathcal{U}(0.2, 0.6)$, and the classifier-free guidance weight is set to 5.  For Stable Diffusion SDS, we sample the timestep $t \sim \mathcal{U}(0.02, 0.5)$, and the classifier-free guidance weight is set to 50. 

The overall $\mathcal{L}_{fine}^{tex}$ loss for the coarse stage can be formulated as a
combination of $\mathcal{L}_{sds}^{z123}$, $\mathcal{L}_{sds}^{sd}$, $\mathcal{L}_{{rgb}}$ and $\mathcal{L}_{{vpc}}$:
\vspace{-3pt}
\begin{equation}
    \begin{aligned}
        \mathcal{L}_{fine}^{tex} = \lambda_1 \mathcal{L}_{sds}^{z123} + \lambda_2 \mathcal{L}_{sds}^{sd} + \lambda_3 \mathcal{L}_{rgb} + \lambda_4 \mathcal{L}_{{vpc}}
    \end{aligned}
\end{equation}
where in practice $\lambda_1=0.002$, $\lambda_2=0.5$, $\lambda_3=10000$, $\lambda_4=10$. We optimize the texture stage using Adam optimizer for 4000 steps with a learning rate 1×10$^{-3}$. To maintain the front/back view details and generate consistent side view texture, we optimize another 2000 steps with $\lambda_1=0$, $\lambda_2=0$, $\lambda_3=10000$, $\lambda_4=100$.

\section*{C. Compare with TeCH}
TeCH\cite{huang2023tech} is our concurrent work,  which is also an optimization-based method that employs \emph{Score Distillation Sampling} during the optimization process. As can be observed in Figure \ref{exp:comp_tech}, TeCH tends to predict a floating human pose and always exhibits a misaligned texture in the hand region. Most importantly, as shown in the back view, TeCH shows an unreasonable texture compared with the input image in terms of texture pattern, texture style, and wrong prediction of the hat in the back head region.

\section*{D. More visual comparison}
\begin{figure*}[htpb]
	\centering
	\includegraphics[width=1.0\linewidth]{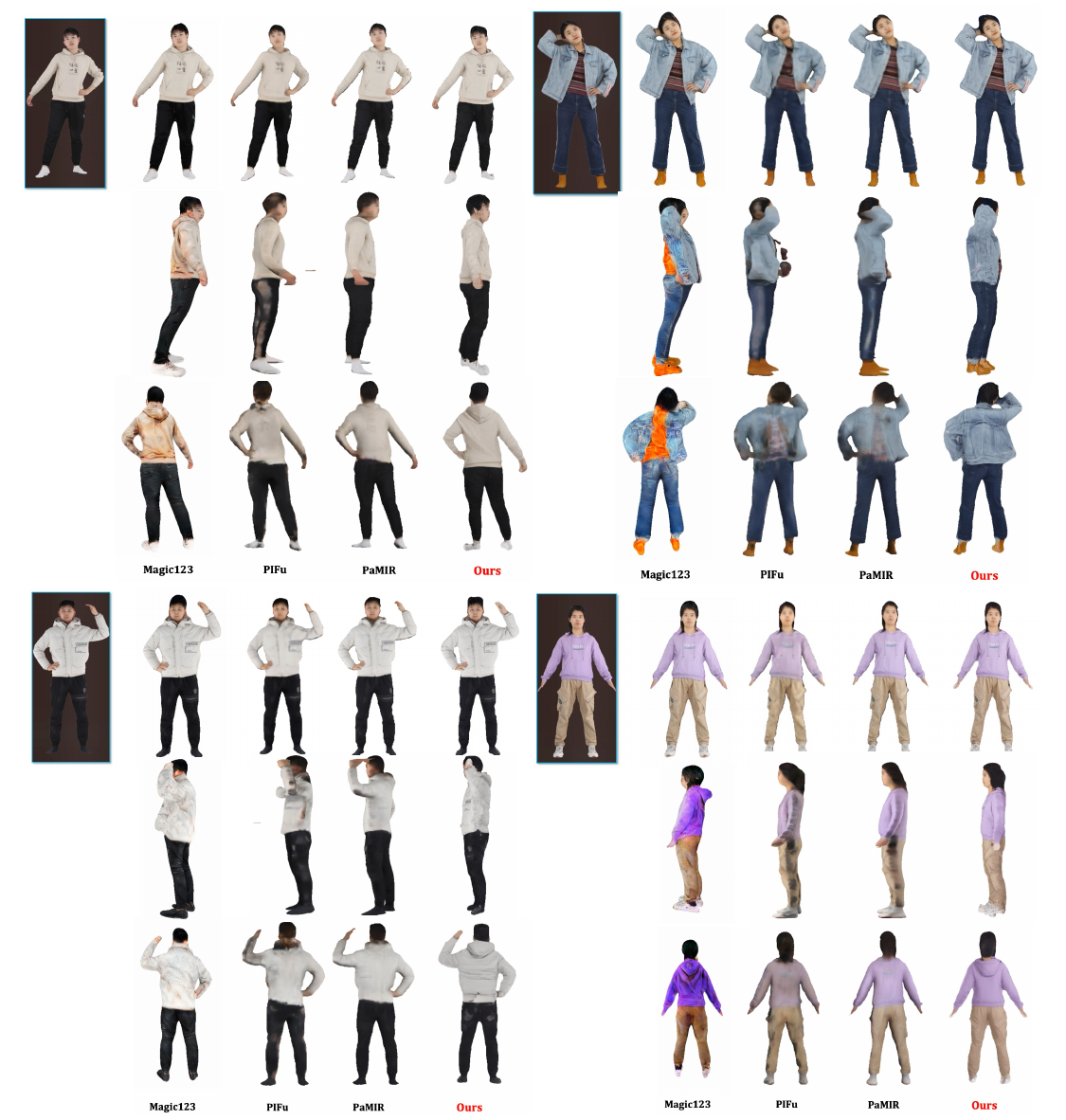}
	\vspace{-16pt}
        \caption{{\textbf{Qualitative comparison results on THuman2.0 dataset.} Compared with them, our methods can render texture-consistent and high-fidelity novel views.}}
        \label{exp:comp_more_thuman}
	\vspace{-10pt}
\end{figure*}

\begin{figure*}[htpb]
	\centering
	\includegraphics[width=1.0\linewidth]{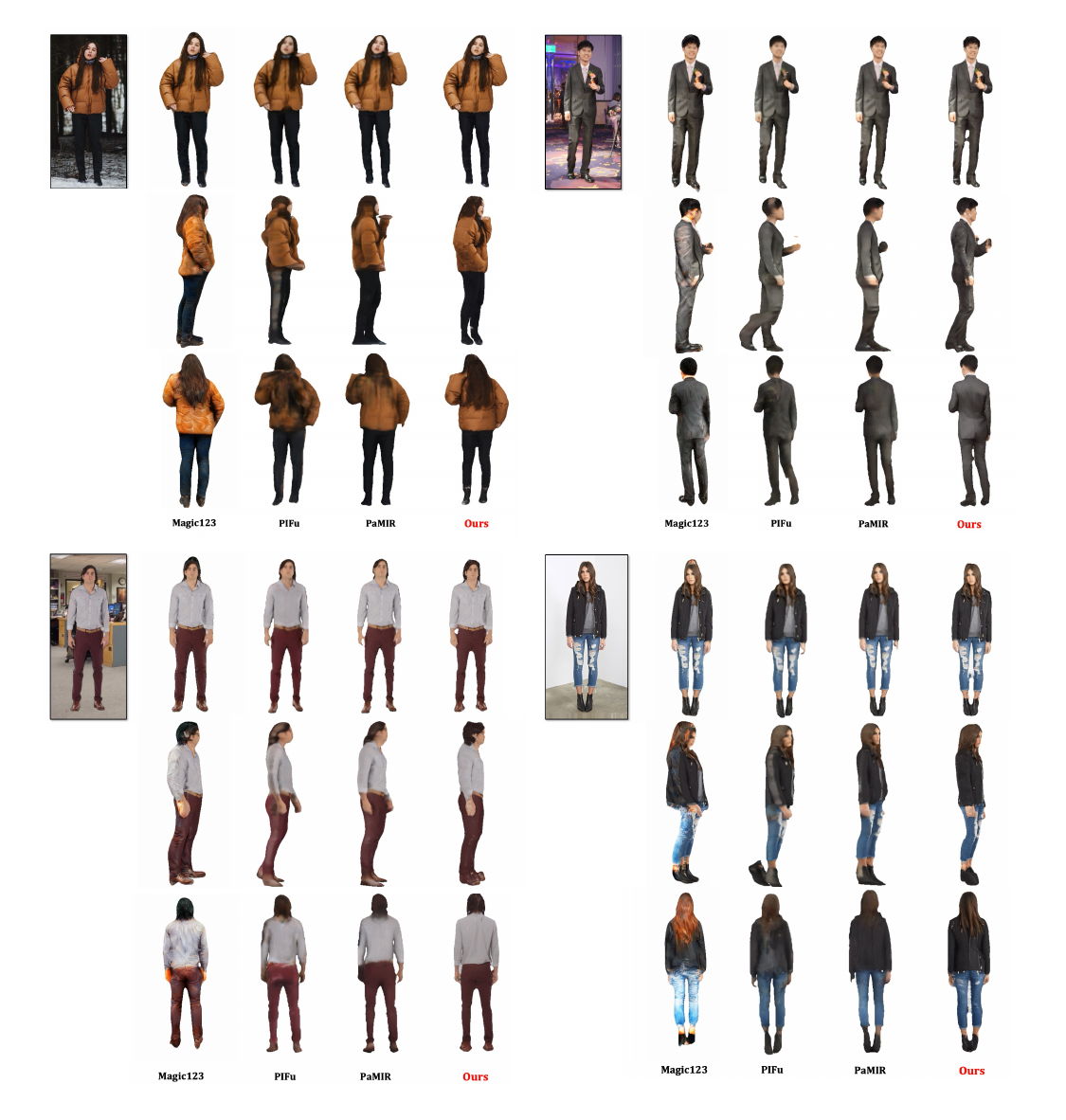}
	\vspace{-16pt}
        \caption{{\textbf{Qualitative results on SSHQ dataset.} Compared with them, our methods can render texture-consistent and high-fidelity novel views.}}
        \label{exp:comp_more_sshq}
	\vspace{-10pt}
\end{figure*}

We provide more visual results in Figure \ref{exp:comp_more_thuman} on THuman2.0 dataset and Figure \ref{exp:comp_more_sshq} on SSHQ dataset. Please Zoom in For more details.

\section*{E. Visual results video}
We also provide two demo videos of free-view human rendering on about 20 human subjects. The first one is a comparison video with PIFu, PaMIR, and Magic123. The second one is a comparison video with TeCH. Each page in the video contains several human cases, we recommend you play the video repeatedly or drag the video progress bar for more details.

\end{document}